\documentclass[pdflatex,sn-mathphys-num]{sn-jnl}

\usepackage{graphicx}%
\usepackage{multirow}%
\usepackage{amsmath,amssymb,amsfonts}%
\usepackage{amsthm}%
\usepackage{mathrsfs}%
\usepackage[title]{appendix}%
\usepackage{xcolor}%
\usepackage{textcomp}%
\usepackage{manyfoot}%
\usepackage{booktabs}%
\usepackage{algorithm}%
\usepackage{algorithmicx}%
\usepackage{algpseudocode}%
\usepackage{listings}%
\usepackage{tikz}
\usetikzlibrary{arrows}

\usepackage{colortbl}
\usepackage{tabularx}
\usepackage{array}
\usepackage{longtable}
\usepackage{caption}
\usepackage{graphicx}

\theoremstyle{thmstyleone}%
\theoremstyle{thmstyletwo}%

\theoremstyle{thmstylethree}%

\begin{document}

\title[From LLMs to multimodal AI]{From large language models to multimodal AI: A scoping review on the potential of generative AI in medicine}

\author*[1]{\fnm{Lukas} \sur{Buess}}\email{lukas.buess@fau.de}
\author[2]{\fnm{Matthias} \sur{Keicher}}
\author[2]{\fnm{Nassir} \sur{Navab}}
\author[1]{\fnm{Andreas} \sur{Maier}}
\author[1]{\fnm{Soroosh} \sur{Tayebi Arasteh}}
\affil[1]{\orgdiv{Pattern Recognition Lab}, \orgname{Friedrich-Alexander-Universität Erlangen-Nürnberg}, \orgaddress{\city{Erlangen}, \country{Germany}}}
\affil[2]{\orgdiv{Computer Aided Medical Procedures}, \orgname{Technical University of Munich}, \orgaddress{\city{Munich}, \country{Germany}}}

\abstract{
Generative artificial intelligence (AI) models, such as diffusion models and OpenAI's ChatGPT, are transforming medicine by enhancing diagnostic accuracy and automating clinical workflows. The field has advanced rapidly, evolving from text-only large language models for tasks such as clinical documentation and decision support to multimodal AI systems capable of integrating diverse data modalities, including imaging, text, and structured data, within a single model. The diverse landscape of these technologies, along with rising interest, highlights the need for a comprehensive review of their applications and potential. This scoping review explores the evolution of multimodal AI, highlighting its methods, applications, datasets, and evaluation in clinical settings. Adhering to PRISMA-ScR guidelines, we systematically queried PubMed, IEEE Xplore, and Web of Science, prioritizing recent studies published up to the end of 2024. After rigorous screening, 144 papers were included, revealing key trends and challenges in this dynamic field. Our findings underscore a shift from unimodal to multimodal approaches, driving innovations in diagnostic support, medical report generation, drug discovery, and conversational AI. However, critical challenges remain, including the integration of heterogeneous data types, improving model interpretability, addressing ethical concerns, and validating AI systems in real-world clinical settings. This review summarizes the current state of the art, identifies critical gaps, and provides insights to guide the development of scalable, trustworthy, and clinically impactful multimodal AI solutions in healthcare.
}
\keywords{Large language models, Generative AI, Multimodal AI, Scoping review}

\maketitle

\section{Introduction}
Generative artificial intelligence (AI), exemplified by models like ChatGPT, has drawn widespread attention for its ability to process and generate human-like text, substantially advancing various domains. In healthcare, these models have rapidly transformed traditional approaches by offering capabilities beyond conventional data analysis \cite{acosta2022multimodal, tayebi2024large}. For instance, large language models (LLMs) have been applied in tasks such as summarizing medical records \cite{cai2021chestxraybert}, assisting in diagnostic reasoning \cite{li2023chatdoctor}, and conducting bioinformatics research \cite{jumper2021highly}. These advancements highlight the ability of LLMs to process and interpret complex clinical language, improving efficiency and accuracy across tasks such as radiology reporting. Recent studies further demonstrate their impact, showing that AI-generated draft radiology reports can reduce reporting time by about 25\% while maintaining diagnostic accuracy \cite{acosta2024impactaiassistanceradiology}, thus addressing workload challenges in clinical practice \cite{van2024adapted}.

However, healthcare data extends far beyond clinical texts, encompassing diverse modalities such as medical images \cite{johnson2019mimic, huang2023inspect}, laboratory results \cite{tayebi2024treasure, khader2023multimodal}, and genomic data \cite{ frankish2021gencode}. To address this diversity, multimodal AI systems have emerged, integrating these data types within a single model. This integration paves the way for comprehensive decision support systems that more closely mimic human clinical reasoning. Recent advancements in multimodal AI represent a significant shift, expanding generative AI applications beyond language-focused tasks to more complex data integration scenarios \cite{zhang2024generalist, hamamci2024foundation, li2024llava}. By unifying text, images, and other clinical data, these systems hold potential for improved diagnostic accuracy and broader applications, from predictive analytics to complex interventional support \cite{ozsoy2024oracle}.

\begin{figure}[h!]
    \centering
    \resizebox{0.82\textwidth}{!}{
        \includegraphics{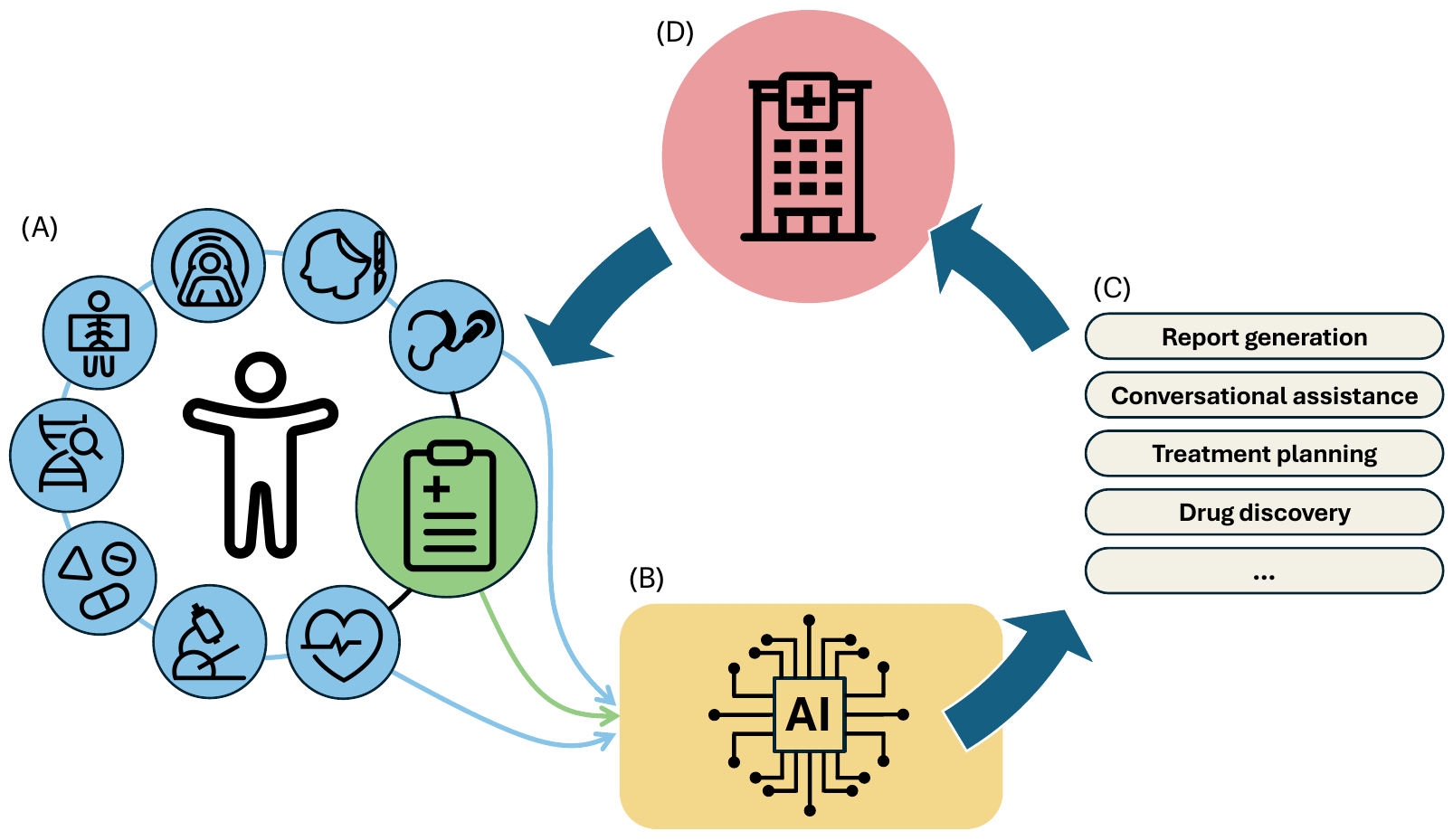}
    }
    \caption{Multimodal AI pipeline in healthcare: (A) Diverse medical data modalities (e.g., images, genomics, and clinical notes) are collected and processed, (B) transformed into unified representations by AI models, (C) used to generate insights such as reports, conversational assistance, and treatment plans, and (D) refined through iterative feedback to continuously optimize data collection and AI performance.}
    \label{fig:overview}
\end{figure}

Several recent review articles have provided valuable overviews of multimodal AI and LLMs. Comprehensive surveys of multimodal large language models (MLLMs) in the broader computer vision domain were presented by \citet{yin2023survey} and \citet{wang2024comprehensive}, highlighting recent advancements, providing a summary of architectural developments, and identifying key trends in model evolution. A broader perspective on multimodal approaches in healthcare was provided by \citet{kline2022multimodal} and \citet{acosta2022multimodal}. \citet{he2024foundation} present a comprehensive collection of foundation models, spanning from image-only architectures to advanced multimodal models.

While previous reviews provide essential insights, the dynamic and rapidly evolving nature of this field necessitates an up-to-date and focused exploration of recent developments in LLM-based multimodal AI for medicine. This review aims to fill this gap by providing a comprehensive overview of the evolution from text-only LLMs to multimodal AI systems in medicine, with a particular emphasis on recent advancements. Unlike prior reviews, we also discuss evaluation methods specifically tailored to the challenges and requirements of medical generative AI, ensuring real-world clinical utility and reliability.

To guide this review, we formulated the following research questions:
\begin{itemize}
    \item What methods are commonly used in the development of generative AI for healthcare applications?
    \item What datasets support the development of generative AI in medical contexts?
    \item Which evaluation metrics are employed to assess the utility of generative AI models in medical contexts?
\end{itemize}

In the following sections, we first outline the methodology employed for literature collection and selection, detailing the search strategy, inclusion criteria, and data extraction processes used to ensure a comprehensive review. We then present our findings, emphasizing the shift from text-only LLMs to multimodal AI systems in medicine, with a particular focus on their applications, datasets, model architectures, and evaluation metrics. Our results reveal a significant shift towards multimodal models, which are driving innovation across various areas of healthcare. However, persistent challenges remain, particularly in the evaluation of these models, including the assessment of their reliability, clinical relevance, and generalizability. Finally, we provide an outlook on the future of generative AI in medicine, offering insights to guide further research and development in this rapidly evolving field.

\section{Methods}
Our scoping review followed the Preferred Reporting Items for Systematic Reviews and Meta-Analyses extension for Scoping Reviews (PRISMA-ScR) \cite{tricco2018prisma, page2021prisma}, which provides a standardized framework for methodological transparency in scoping reviews. This section details the data collection methods used in our review. The complete PRISMA-ScR checklist is available in Supplementary Table \ref{tab:checklist}.

\subsection{Eligibility criteria}
We included studies published between January 2020 and December 2024 to capture recent advancements in the rapidly advancing field of generative AI in medicine. Only original research in English was eligible, as our focus is on primary contributions rather than synthesized findings. Review and meta-analysis papers were therefore excluded. We included peer-reviewed conference and journal publications, alongside manually selected preprints with high relevance and potential impact. To ensure a comprehensive overview, foundational dataset papers published before 2020 were also included when they were widely used in the selected studies or remained relevant for benchmarking. This approach ensured a focus on current, state-of-the-art developments in multimodal AI applications in medicine.

\subsection{Information sources}
We performed a systematic search in PubMed, IEEE Xplore, and Web of Science, employing a standardized set of keywords derived from our research objectives. Full search queries are detailed in Supplementary Table \ref{tab:queries}. The searches, conducted on October 1, 2024, were imported into Rayyan \cite{ouzzani2016rayyan}, a web-based tool designed to facilitate literature screening and semi-automated duplicate removal.

\subsection{Search strategy}
The literature search consisted of a systematic database search structured into two subsearches to capture the development and application of text-only LLMs and multimodal models in medicine. The first subsearch targeted text-only LLMs using the keyword groups "medical" and "language model". The second subsearch focused on multimodal models, using three groups of keywords: "medical", "language model", and "multimodal". The full search queries, including the specific combinations used, are provided in Supplementary Table \ref{tab:queries}. Additionally, a manual search was performed to identify recent preprints, datasets, and other resources not captured by the database search, which continued through the end of 2024 to ensure the inclusion of the most current and impactful studies.

\subsection{Inclusion and exclusion criteria}
The selection process began with structured database queries, followed by duplicate removal, title and abstract screening, and subsequent full-text reviews for potentially relevant papers. We excluded articles that were non-medical or lacked methodological novelty. To ensure balanced representation across application areas, we aimed for proportional inclusion from prevalent fields, such as X-ray report generation.

\subsection{Synthesis of results}
The selected papers were categorized through a two-step process. First, they were grouped by topics, including text-only LLMs, multimodal models, datasets, and evaluation metrics. Within each topic, papers were further categorized based on their application areas. This dual-layer categorization provides a structured overview of developments in generative AI for medicine, illustrating the progression from text-only LLMs to multimodal models. Key publications are summarized through narrative descriptions and tables, offering insights into methodological approaches, application domains, datasets, and evaluation frameworks to provide a comprehensive understanding of current trends and challenges. Tables \ref{tab:llmmethods} (text-only LLMs), \ref{tab:llmdatasets} (text-only datasets), \ref{tab:clip} (contrastive learning methods), \ref{tab:mllms} (MLLMs), \ref{tab:multimodaldatasets} (multimodal datasets), and \ref{tab:evaluationmetrics} (evaluation metrics) summarize the results.

\section{Included studies}
\begin{figure}[h!]
    \centering
    \resizebox{1.0\textwidth}{!}{
        \includegraphics{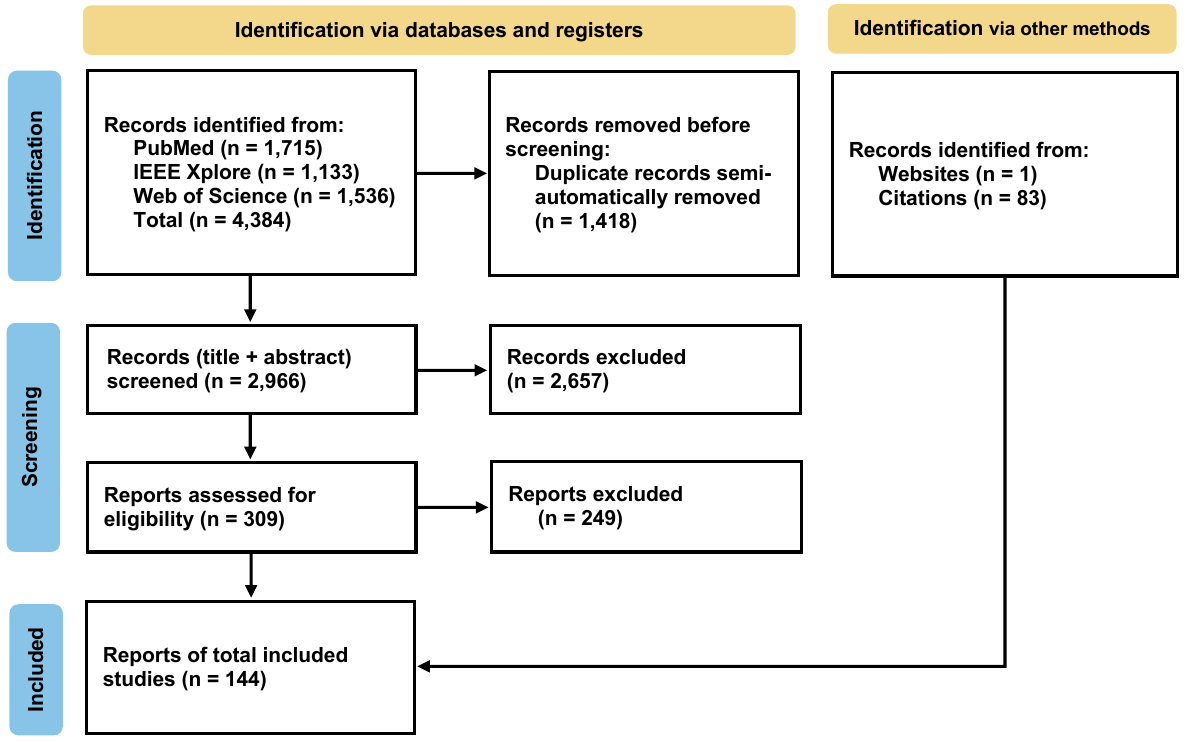}
    }
    \caption{PRISMA flow diagram illustrating the study selection process for the scoping review. The diagram shows the number of records identified through database searches and manual searches, the removal of duplicates, the screening of titles and abstracts, the review of full-text articles, and the final inclusion of studies in the review.}
    \label{fig:flowdiagram}
\end{figure}

A total of 4,384 papers were retrieved from three databases. After removing duplicates, 2,656 articles were excluded during the initial screening based on their titles and abstracts, following the predefined inclusion and exclusion criteria. The remaining articles underwent a full-text review, during which both relevance and topic diversity were considered to avoid overrepresentation of similar studies. This step led to the exclusion of an additional 249 papers. Ultimately, 60 papers from the database search were included in the review. Additionally, 83 papers were identified through manual searches to capture the most current and relevant studies not covered in the database queries. Figure \ref{fig:flowdiagram} provides an overview of the full screening process. In total, 144 papers were included in this review.

\section{Language models in medicine}
Mono-modal LLMs, which process textual data exclusively, have laid the foundation for the development of multimodal systems, demonstrating remarkable capabilities in understanding and generating human-like text. In the medical domain, LLMs demonstrated high effectiveness in processing and analyzing complex clinical data, enabling advancements in applications such as clinical documentation, medical literature summarization, and diagnostic support \cite{cai2021chestxraybert, jiang2023health}. Their success is based on the transformer architecture, introduced in the landmark paper “Attention Is All You Need” \cite{vaswani2017attention}, which employs self-attention mechanisms to effectively capture contextual relationships and long-range dependencies in text. This architecture has enabled LLMs to scale effectively, making them capable of processing medical texts.

\subsection{LLM methods}
LLMs tailored to medical applications (Table \ref{tab:llmmethods}) leverage various approaches to adapt general-purpose models for specialized medical tasks. A prevalent method is supervised finetuning (SFT), where general LLMs are finetuned on domain-specific datasets, such as biomedical literature and clinical notes, to enhance their understanding of medical concepts and vocabulary. This approach has been instrumental in models like BioBERT and BioMistral, which adapt general-purpose language models for biomedical applications \cite{lee2020biobert, labrak2024biomistral}.

In contrast to SFT, prompt engineering techniques have emerged as a lightweight alternative for guiding pretrained models without additional training, relying on carefully designed input prompts to achieve strong task performance in medical text understanding and generation \cite{wang2024prompt}.

Advanced alignment techniques such as reinforcement learning from human feedback (RLHF) have been developed to further refine the outputs of LLMs for medical applications. RLHF leverages reward models trained on expert feedback to align model responses with clinical expectations. However, due to the cost of obtaining expert feedback in the healthcare domain, reinforcement learning from AI feedback (RLAIF) has emerged as an alternative \cite{lee2023rlaif}. This technique replaces human feedback with evaluations from auxiliary AI models, reducing reliance on scarce human resources while maintaining alignment capabilities. A notable example is HuatuoGPT \cite{zhang2023huatuogpt}, which uses RLAIF for clinical alignment.

Another recent development in model adaption is chain-of-thought (CoT) prompting, a technique where models generate intermediate reasoning steps before producing a final answer. By breaking down complex tasks into substeps, CoT enhances model explainability and task performance, which is especially valuable in the medical domain as it not only improves accuracy but also increases trust in the model’s reasoning process. For example, HuatuoGPT-o1 \cite{chen2024huatuogpto1} applies CoT prompting to improve medical response clarity and ensure step-by-step diagnostic reasoning.

An additional adaptation technique is retrieval augmented generation (RAG) \cite{lewis2020retrieval}, which equips LLMs with mechanisms to query external knowledge bases during inference. This approach enables models to access up-to-date information, such as medical guidelines or recent research findings, without requiring retraining. For instance, Almanac \cite{zakka2024almanac}, ChatDoctor \cite{li2023chatdoctor}, and RadioRAG \cite{arasteh2024radioragfactuallargelanguage} combine generative capabilities with retrieval systems. However, maintaining the retrieval database and ensuring its comprehensiveness pose ongoing challenges \cite{li2023chatdoctor, gilbert2024augmented}.

\begin{table}[b]
    \centering
    \resizebox{0.6\textwidth}{!}{
        \begin{tabular}{p{3.8cm} >{\centering\arraybackslash}p{4.5cm}}
        \toprule
        \textbf{Study} & \textbf{Downstream task} \\
        \midrule
        
        \multicolumn{2}{l}{\textbf{Clinical text}} \\
        \midrule
        Almanac \cite{zakka2024almanac} & QA \\
        BioALBERT \cite{naseem2021bioalbert} & NER \\
        BioBERT \cite{lee2020biobert} & NER, QA \\
        BioGPT \cite{luo2022biogpt} & Classification, QA \\
        BioMistral \cite{labrak2024biomistral} & QA \\
        ChatDoctor \cite{li2023chatdoctor} & Dialogue \\
        ChestXRayBERT \cite{cai2021chestxraybert} & Summarization \\
        DRG-LLaMA \cite{wang2024drg} & Classification \\
        GatorTron \cite{yang2022large} & QA \\
        HuatuoGPT \cite{zhang2023huatuogpt} & Dialogue \\
        HuatuoGPT-o1 \cite{zhang2023huatuogpt} & Dialogue \\
        \citet{johnson2020deidentification} & Deidentification \\
        \citet{kresevic2024optimization} & Summarization \\
        \citet{mahendran2021extracting} & NER \\
        MAPLEZ \cite{lanfredi2025enhancing} & Classification \\
        Med-BERT \cite{liu2021med} & NER \\        
        MedAlpaca \cite{han2023medalpaca} & QA \\
        MEDITRON-70B \cite{chen2023meditron} & QA \\
        MED-PaLM \cite{singhal2023large} & QA \\
        MMed-Llama 3 \cite{qiu2024towards} & QA \\
        \citet{mu2021bert} & Classification \\
        NYUTron \cite{jiang2023health} & Clinical outcome prediction \\
        PMC-LLaMA \cite{wu2024pmc} & QA \\
        PodGPT \cite{jia2024medpodgpt} & QA \\
        RadBERT \cite{yan2022radbert} & Classification, Summarization \\
        \citet{schmidt2024generative} & Error detection \\
        
        \midrule
        \multicolumn{2}{l}{\textbf{Bioinformatics}} \\
        \midrule
        AlphaFold \cite{jumper2021highly} & Structure prediction \\
        BioPhi \cite{prihoda2022biophi} & Antibody design \\
        CADD v1.7 \cite{schubach2024cadd} & Scoring \\
        DNABERT \cite{ji2021dnabert} & Structure analysis \\
        Geneformer \cite{theodoris2023transfer} & Classification \\
        \citet{hie2024efficient} & Antibody design \\
        MSA Transformer \cite{rao2021msa} & Structure analysis \\
        ProGen \cite{madani2023large} & Structure prediction \\
        ProtGPT2 \cite{ferruz2022protgpt2} & Protein design \\
        ProtTrans \cite{elnaggar2021prottrans} & Structure analysis \\
        scBERT \cite{yang2022scbert} & Classification \\
        ToxinPred 3.0 \cite{rathore2024toxinpred} & Classification \\
        
        \bottomrule
    \end{tabular}
    }
    \captionsetup{width=0.6\linewidth}
    \caption{Summary of LLM methods, categorized by their application to clinical text and bioinformatics tasks. The table includes method names and target applications. (Abbreviations: NER - named entity recognition, QA - question answering)}
    \label{tab:llmmethods}
\end{table}

\subsection{LLM applications}
LLMs have revolutionized various applications in biomedical language processing, demonstrating utility across a range of tasks. In named entity recognition (NER), they enable the extraction of critical medical entities, such as diseases, drugs, and symptoms from unstructured text. This capability supports clinical data annotation, which is crucial for automated clinical decision support systems \cite{naseem2021bioalbert}.

Dialogue systems represent another application of LLMs in medicine. Models like ChatDoctor \cite{li2023chatdoctor} and HuatuoGPT \cite{zhang2023huatuogpt} facilitate patient interactions, simulate doctor-patient consultations, and assist in providing medical information and guidance. These systems aim to reduce barriers to medical access by providing instant responses.

In summarization tasks, medical LLMs condense lengthy electronic health records (EHRs) into concise summaries. This application significantly reduces the documentation burden on healthcare providers and aids decision-making by presenting critical patient information in a structured format \cite{cai2021chestxraybert, nowak2023transformer}.

Deidentification and privacy-preserving applications are critical areas where LLMs contribute to medical data management by safeguarding patient confidentiality in sensitive clinical texts. LLMs can automate the removal of protected health information from medical documents by anonymizing identifiers such as names and dates while preserving data utility \cite{johnson2020deidentification, lanfredi2025enhancing}.

Text classification tasks also benefit from LLM advancements, with applications such as predicting patient outcomes and categorizing medical literature \cite{jiang2023health}.

In bioinformatics, LLMs have expanded beyond language processing to analyze biological sequences like DNA, RNA, and proteins. Models such as DNABERT \cite{ji2021dnabert} have advanced gene annotation, while AlphaFold \cite{jumper2021highly} has achieved groundbreaking success in protein structure prediction.

\subsection{LLM datasets}
The development of medical LLMs relies on diverse and specialized datasets that capture the complexity of medical language, context, and tasks. These datasets fall into categories such as clinical text, domain-specific literature, conversational data, and bioinformatics resources, each serving distinct purposes in the development of medical LLMs. These datasets enable general-purpose LLMs to align with the medical domain, which is critical for achieving reliable and accurate outputs in clinical settings.

Clinical text datasets play a central role in training medical LLMs (see Table \ref{tab:llmdatasets}). For instance, EHR datasets like MIMIC-IV \cite{johnson2023mimic} provide a rich source of structured and unstructured clinical data, commonly used for tasks such as summarization and NER, which are both essential for automating documentation and decision-making processes in healthcare. The eICU-CRD dataset \cite{pollard2018eicu}, another EHR resource, focuses on intensive care unit patient data, further broadening the scope of potential applications.

To introduce domain-specific knowledge into LLMs, datasets like GAP-Replay \cite{chen2023meditron} and MedC-K \cite{wu2024pmc}, composed of biomedical literature and textbooks, are essential. These datasets are designed to equip models with the specialized terminology and reasoning patterns found in biomedical research and education.

For conversational AI in medicine, dialogue datasets are crucial. MedDialog \cite{zeng2020meddialog} provides examples of doctor-patient interactions, enabling LLMs to learn medical dialogues, including patient concerns, physician responses, and diagnostic reasoning. These datasets are essential for developing medical conversational assistance systems capable of simulating clinical dialogues and supporting in patient education, diagnostic reasoning, and post-treatment follow-ups.

Bioinformatics datasets extend the scope of LLM applications beyond clinical text, supporting tasks in genomics and molecular biology. Resources like AlphaFold DB \cite{jumper2021highly} and UniProtKB \cite{uniprot2023uniprot} provide structured data for protein structure and sequence analysis, making them valuable for drug discovery and molecular research. Similarly, genomic datasets such as GENCODE \cite{frankish2021gencode} and GenBank \cite{benson2012genbank} offer data for tasks like gene prediction, helping models to better understand complex biological patterns.

\begin{table}[h]
    \centering
    \begin{tabular}{p{3.5cm} >{\centering\arraybackslash}p{4.0cm} >{\centering\arraybackslash}p{3.5cm}}
        \toprule
        \textbf{Dataset} & \textbf{Size} & \textbf{Application} \\
        \midrule

        \multicolumn{2}{l}{\textbf{Clinical text}} \\
        \midrule
        eICU-CRD \cite{pollard2018eicu} & 200K instances & EHR \\
        GAP-Replay \cite{chen2023meditron} & 48.1B tokens & Literature \\
        MedDialog-EN \cite{zeng2020meddialog} & 250K dialogues & Dialogue \\
        MedC-K \cite{wu2024pmc} & 4.8M papers, 30K textbooks & Literature \\
        MedC-I \cite{wu2024pmc} & 202M tokens & Dialogue, QA \\
        Medical Meadow \cite{han2023medalpaca} & 160K instances & QA \\
        MIMIC-IV \cite{johnson2023mimic} & 299K patients & EHR \\
        MMedC \cite{qiu2024towards} & 25.5B tokens & Multilingual literature \\
        MultiMedQA \cite{singhal2023large} & 213K instances & QA \\

        \midrule
        \multicolumn{2}{l}{\textbf{Bioinformatics}} \\
        \midrule
        AlphaFold DB \cite{jumper2021highly} & 200M entries & Protein Design \\
        CPTAC Data Portal \cite{edwards2015cptac} & NA & Genomics, Protein Design \\
        GenBank \cite{benson2012genbank} & NA sequences & Genomics \\
        GENCODE \cite{frankish2021gencode} & NA & Genomics \\
        UniProtKB \cite{uniprot2023uniprot} & 227M sequences & Protein Design \\
        
        \bottomrule
    \end{tabular}
    \caption{Summary of datasets used for training medical LLMs, categorized into clinical text and bioinformatics data. The table includes dataset names, sizes, and primary application areas. (Abbreviations: NA - not available, NER - named entity recognition, QA - question answering, EHR - electronic health record)}
    \label{tab:llmdatasets}
\end{table}

\section{Multimodal language models in medicine}
By showcasing the potential of LLMs in processing clinical text, these models have established a strong foundation for integrating additional modalities, leading to the development of multimodal language models specifically designed for healthcare. Multimodal models combine diverse data types, such as text and medical images, to tackle complex medical tasks, including report generation \cite{bannur2024maira, pellegrini2023radialog}, image-text retrieval \cite{tiu2022expert, endo2021retrieval}, and medical consultation \cite{hamamci2024foundation}. By building on advancements in LLMs, multimodal language models improve the integration and contextual understanding of multimodal medical data. This section provides an overview of recent architectures and methods addressing the unique challenges posed by multimodal medical data.

\subsection{Architectures}
Before presenting the literature, we briefly outline the two primary architectures in multimodal AI, i.e., the contrastive language-image pretraining (CLIP) and MLLMs (see Figure \ref{fig:architectures}). These architectures serve as foundational frameworks for integrating multiple data modalities in medical AI. Although CLIP \cite{radford2021learning} is not inherently generative, its ability to align images and text within a shared embedding space makes it a crucial component in multimodal AI systems.

\begin{figure}[h!]
    \centering
    \resizebox{\textwidth}{!}{
        \includegraphics{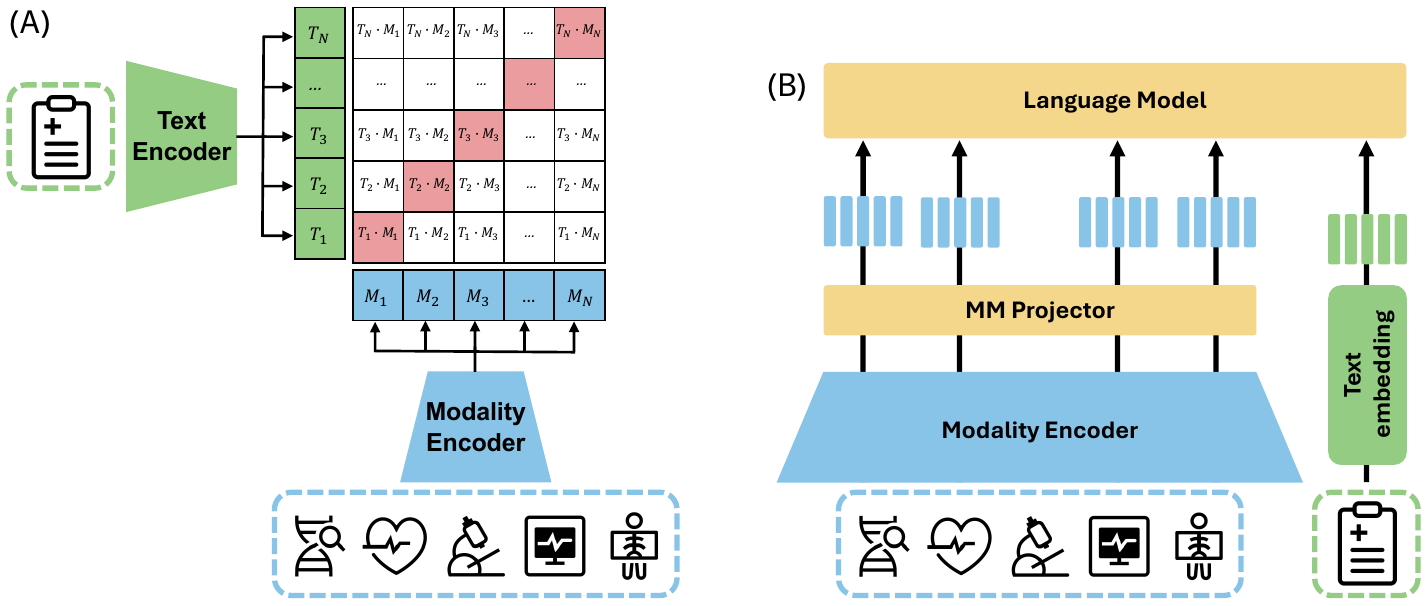}
    }
    \caption{Multimodal architectures: (A) CLIP-based models, which align embeddings of different modalities in a shared latent space, and (B) LLM-based models, which directly integrate different modality inputs through feature extraction and projection into LLM's ebbedding space.}
    \label{fig:architectures}
\end{figure}

CLIP \cite{radford2021learning} is designed to align different modalities, such as image and text, in a shared embedding space. Although originally developed for image-text pairs, its framework can be extended to other modalities, making it a versatile tool for various multimodal learning tasks. By jointly training on paired modalities data, it excels in tasks like zero-shot image classification \cite{tiu2022expert, blankemeier2024merlin}, where new classes can be recognized without additional training. This makes CLIP particularly useful for situations where annotated medical data is limited.

On the other hand, MLLMs, such as LLaVA \cite{liu2024visual}, extend the capabilities of LLMs by integrating non-textual data directly into their embeddings. This integration allows for a more holistic understanding of complex datasets, combining linguistic context with multimodal features like images or clinical measurements. These models excel in tasks such as radiology report generation \cite{bannur2024maira, pellegrini2023radialog}, question answering (QA) about medical images \cite{tu2024towards, zhou2024generalist}, and decision support in diagnosis \cite{zhang2024generalist, blankemeier2024merlin, wu2023towards}.

By leveraging complementary strengths, these architectures address the diverse challenges posed by multimodal medical data. CLIP is effective for aligning different data modalities, while MLLMs excel in diagnostic reasoning, together forming a powerful combination for improving multimodal AI in medicine.

\subsection{Multimodal LLM methods}
Modality alignment is a fundamental step for most MLLMs. Many approaches leverage CLIP-based methods (Table \ref{tab:clip}), which primarily focus on learning a shared latent space where modalities can be jointly represented for downstream tasks.

\begin{table}[b]
    \centering
    \resizebox{1.0\textwidth}{!}{
        \begin{tabular}{p{3cm} >{\centering\arraybackslash}p{5.5cm} >{\centering\arraybackslash}p{7cm}}
        \toprule
        \textbf{Study} & \textbf{Modalities} & \textbf{Application} \\
        \midrule
        BiomedCLIP \cite{zhang2024multimodal} & Medical images, Descriptions & Classification, Retrieval, Visual QA \\
        BioViL \cite{boecking2022making} & X-ray, Reports & Classification, Grounding \\
        BioViL-T \cite{bannur2023learning} & X-ray, Reports & Classification, Grounding, Reporting \\
        CheXzero \cite{tiu2022expert} & X-ray, Reports & Classification, Retrieval \\
        ConVIRT \cite{zhang2022contrastive} & X-ray, Reports & Classification, Retrieval \\
        CPLIP \cite{javed2024cplip} & Histopathology images, Descriptions & Classification \\
        CT-CLIP \cite{hamamci2024foundation} & CT, Reports, Labels & Classification, Retrieval \\
        CT Foundation \cite{yang2024advancing} & CT, Reports & Classification, Retrieval \\
        CXR-RePaiR \cite{endo2021retrieval} & X-ray, Reports & Reporting \\
        ETP \cite{liu2024etp} & ECG, Reports & Classification \\
        FairCLIP \cite{luo2024fairclip} & SLO fundus images, Clinical notes & Classification \\
        FiVE \cite{li2024generalizable} & Histopathology images, Descriptions & Classification \\
        FlexR \cite{keicher2024flexr} & X-ray, Reports & Classification \\
        GLoRIA \cite{huang2021gloria} & X-ray, Reports & Classification, Retrieval, Segmentation \\
        KAD \cite{zhang2023knowledge} & X-ray, Reports & Classification \\
        MaCo \cite{huang2024enhancing} & X-ray, Reports & Classification \\
        MCPL \cite{wang2024mcpl} & X-ray, Reports & Classification \\
        MedImageInsight \cite{codella2024medimageinsight} & Medical images, Descriptions & Classification, Retrieval, Reporting \\
        Med-MLLM \cite{liu2023medical} & CT, X-ray, Descriptions & Classification, Reporting \\
        Merlin \cite{blankemeier2024merlin} & CT, EHR, Reports & Classification, Retrieval, Reporting, Segmentation \\
        MedViLL \cite{moon2022multi} & X-ray, Reports & Classification, Retrieval, Reporting, Visual QA \\
        MoleculeSTM \cite{liu2023multi} & Molecule structure, Descriptions & Retrieval \\
        MolLM \cite{tang2024mollm} & Molecule structures, Descriptions & Retrieval, Molecule description \\
        PLIP \cite{huang2023visual} & Histopathology images, Descriptions & Classification, Retrieval \\
        Prov-GigaPath \cite{xu2024whole} & Histopathology images, Reports & Classification \\
        UniMed-CLIP \cite{khattak2024unimed} & Medical images, Captions & Classification \\
        Xplainer \cite{pellegrini2023xplainer} & X-ray, Reports & Classification \\
        \bottomrule
    \end{tabular}
    }
    \captionsetup{width=1.0\linewidth}
    \caption{Summary of multimodal CLIP-based methods. The table includes method names, the modalities utilized (e.g., text and medical images), and the primary tasks addressed, such as image-text retrieval, report generation, and disease classification. (Abbreviations: QA - question answering)}
    \label{tab:clip}
\end{table}

For instance, BiomedCLIP \cite{zhang2024multimodal} uses contrastive learning to align medical images with paired reports, achieving state-of-the-art results in retrieval tasks. Building on this framework, CheXzero \cite{tiu2022expert} adapts CLIP for zero-shot classification of X-ray images, while CT-CLIP \cite{hamamci2024foundation} extends this approach to computed tomography (CT) scans. Similarly, UniMed-CLIP \cite{khattak2024unimed} enhances this paradigm by using classification datasets augmented by LLM-generated captions to train a foundation model capable of handling various medical image modalities.

More recent efforts have focused on large-scale pretrained models developed by industry leaders, aiming to generalize across diverse medical imaging tasks. Models like CT Foundation \cite{yang2024advancing} and MedImageInsight \cite{codella2024medimageinsight}, accessible via application programming interfaces (APIs), exemplify this trend by offering robust pretrained embeddings that address data scarcity in medical imaging and support downstream applications.

While many CLIP-based methods focus on aligning text with medical images, recent approaches have extended this to other modalities. For example, ETP \cite{liu2024etp} aligns electrocardiogram (ECG) signals \cite{ran2024joint, kang2023gan} with clinical reports, while MolLM \cite{tang2024mollm} pairs chemical structures with textual descriptions to support drug discovery.

\begin{table}[h!]
    \centering
    \resizebox{1.0\textwidth}{!}{ 
        \begin{tabular}{p{3.3cm} >{\centering\arraybackslash}p{5.5cm} >{\centering\arraybackslash}p{7cm}}
        \toprule
        \textbf{Study} & \textbf{Modalities} & \textbf{Downstream task} \\
        \midrule
        \citet{alsharid2022gaze} & US video, Transcriptions, Gaze data & Captioning \\
        AutoRG-Brain \cite{lei2024autorg} & MRI, Reports, Masks & Reporting, Grounding \\
        BiomedGPT \cite{zhang2024generalist} & Medical images, Literature, EHR & Reporting, Summarization, Visual QA \\
        BioMed-VITAL \cite{cui2024biomedical} & Medical images, Instructions & Visual QA \\
        ChatCAD \cite{wang2023chatcad} & X-ray, Reports & Reporting \\
        CheXagent \cite{chen2024chexagent} & X-ray, Reports & Classification, Reporting, Grounding \\
        COMG \cite{gu2024complex} & X-ray, Reports, Masks & Reporting \\
        CT-CHAT \cite{hamamci2024foundation} & CT, Reports & Reporting, Visual QA \\
        FFA-GPT \cite{chen2024ffa} & Fundus fluorescein angiography, Reports & Reporting, Visual QA \\
        GenerateCT \cite{hamamci2023generatect} & CT, Reports & Image generation \\
        \citet{huh2023improving} & X-ray, Reports & Reporting \\
        LLaVA-Med \cite{li2024llava} & Medical images, Captions & Visual QA \\
        LViT \cite{li2023lvit} & CT, X-ray, Masks, Text annotations & Segmentation \\
        M3D-LaMed \cite{bai2024m3d} & CT, Reports, Masks & Reporting, Visual QA, Segmentation \\
        MAIRA-2 \cite{bannur2024maira} & X-ray, Reports, Masks & Reporting, Grounding \\
        MAIRA-Seg \cite{sharma2024maira} & X-ray, Reports, Masks & Reporting \\
        Med-Flamingo \cite{moor2023med} & Medical images, Captions & Visual QA \\
        Med-PaLM M \cite{tu2024towards} & Medical images, Reports, Genomics & Classification, Reporting, Visual QA, Summarization \\
        MedVersa \cite{zhou2024generalist} & CT, X-ray, Dermatology images, Reports & Classification, Reporting, Visual QA, Segmentation \\
        MMBERT \cite{khare2021mmbert} & Radiology images, Captions & Visual QA \\
        MVG \cite{cao2024medical} & Medical images, Text & Disease simulation \\
        ORacle \cite{ozsoy2024oracle} & Multi-view images, SSG, Descriptions & OR scene graph prediction \\
        PathChat \cite{lu2024multimodal} & Histopathology images, QA-pairs & Visual QA \\
        PathLDM \cite{yellapragada2024pathldm} & Histopathology images, Reports & Image generation \\  
        QUILT-LLaVA \cite{seyfioglu2024quilt} & Histopathology images, QA-pairs & Visual QA \\
        R2GenGPT \cite{wang2023r2gengpt} & X-ray, Reports & Reporting \\
        RaDialog \cite{pellegrini2023radialog} & X-ray, Reports & Reporting, Dialogue \\
        RadFM \cite{wu2023towards} & Medical images, Reports, Descriptions & Reporting, Visual QA \\
        ReXplain \cite{luo2024rexplain} & Video, Reports, Masks & Video report generation \\
        RGRG \cite{tanida2023interactive} & X-ray, Reports, Bounding-boxes & Reporting \\
        RoentGen \cite{bluethgen2024vision} & X-ray, Reports & Image generation \\
        SkinGPT-4 \cite{zhou2024pre} & Dermatology images, Clinical notes & Visual QA, Dialogue \\
        Surgical-VQLA++ \cite{bai2025surgical} & Surgical images, QA-pairs & Visual QA \\
        Universal Model \cite{liu2023clip} & CT, Masks, Descriptions & Segmentation \\
        Vote-MI \cite{wangenhancing} & MRI, Reports & Visual QA \\   
        \bottomrule
    \end{tabular}
    }
    \captionsetup{width=1.0\linewidth}
    \caption{Summary of multimodal MLLM-based methods. The table includes method names, the modalities utilized (e.g., text and medical images), and the primary tasks addressed, such as report generation, visual QA, and disease classification. (Abbreviations: QA - question answering)}
    \label{tab:mllms}
\end{table}

LLM-based methods, in contrast to CLIP approaches, directly integrate multimodal inputs into the language model’s embeddings, enabling more complex reasoning and generative tasks. These approaches rely on modality-specific encoders to process non-textual data, converting them into feature embeddings compatible with the LLM’s text-based representation space (Table \ref{tab:mllms}). For instance, SkinGPT-4 \cite{zhou2024pre} and RaDialog \cite{pellegrini2023radialog} integrate features from two-dimensional (2D) images, while models like Merlin \cite{blankemeier2024merlin} and CT-CHAT \cite{hamamci2024foundation} extend this capability to volumetric three-dimensional (3D) CT data. Some models, such as MAIRA-2 \cite{bannur2024maira} and AutoRG-Brain \cite{lei2024autorg}, further ground text predictions by incorporating bounding boxes and segmentation masks, enabling interactive, region-based report generation for enhanced explainability \cite{tanida2023interactive}.

Current advancements also focus on text-guided segmentation and synthetic medical image generation. Text-guided segmentation models like LViT create segmentation masks from textual prompts, enabling tasks such as tumor detection and organ identification \cite{li2023lvit}. Beyond segmentation, synthetic image generation has emerged as another multimodal approach for data augmentation and model training. Methods such as GenerateCT \cite{hamamci2023generatect} for CT volumes and RoentGen \cite{bluethgen2024vision} for X-rays use text-conditioned diffusion models to produce realistic medical images \cite{khader2023denoising}.

Generalist models, such as BiomedGPT \cite{zhang2024generalist} and MedVersa \cite{zhou2024generalist}, unify multiple modalities and tasks through shared representations or mixture-of-experts strategies. These models employ specialized modules to process different modalities while a central language model coordinates their outputs, enabling tasks such as classification, segmentation, retrieval, and visual QA. This approach highlights the scalability and versatility of generalist models in addressing complex multimodal challenges in medicine.

\subsection{Multimodal LLM applications}
MLLMs have been increasingly applied across diverse medical tasks, showcasing their potential to transform clinical workflows and decision support systems. This section highlights key applications where MLLMs contribute to improving healthcare.

A key advancement in multimodal AI is generalist models capable of handling diverse medical data types and tasks. Models such as BiomedGPT \cite{zhang2024generalist} and RadFM \cite{wu2023towards} support a wide range of imaging modalities and anatomical regions, enabling comprehensive diagnostic assistance across multiple specialties.

Radiology report generation remains one of the most important applications of MLLMs in healthcare, providing detailed textual descriptions directly from medical images. Systems such as MAIRA-2 \cite{bannur2024maira} and RaDialog \cite{pellegrini2023radialog} have demonstrated their ability to generate comprehensive reports from X-rays, while CT-CHAT \cite{hamamci2024foundation} and AutoRG-Brain \cite{lei2024autorg} extend this capability to CT and magnetic resonance imaging (MRI) scans, respectively. These tools assist radiologists by automating preliminary reporting and standardizing documentation, potentially reducing reporting delays.

Visual QA systems support clinicians in querying medical images using natural language prompts, supporting real-time decision-making and diagnostic interpretation. For instance, models like LLaVA-Med \cite{li2024llava} and Med-Flamingo \cite{moor2023med} provide concise, contextually relevant answers to clinical queries, assisting radiologists and physicians in complex cases.

Synthetic medical image generation has become increasingly important for data augmentation and simulating rare pathological conditions. Models like GenerateCT \cite{hamamci2023generatect} and RoentGen \cite{bluethgen2024vision} generate realistic CT and X-ray images from textual prompts, enhancing dataset diversity.

Semantic scene modeling is another emerging application where models create structured representations of complex environments, such as the operating room. For example, ORacle \cite{ozsoy2024oracle} generates semantic scene graphs to assist with surgical planning and intraoperative navigation by representing tools, anatomy, and procedural stages in a comprehensive framework.

Finally, systems like ReXplain \cite{luo2024rexplain} aim to bridge communication gaps between clinicians and patients. By transforming radiology reports into patient-friendly video summaries, these models provide an accessible way to convey complex clinical information, further highlighting multimodal AI’s potential to improve patient care.

\subsection{Multimodal LLM datasets}
Multimodal datasets integrating images, text, and other clinical information (Table \ref{tab:multimodaldatasets}) are essential for tasks such as radiology report generation, visual QA, and cross-modal retrieval. These datasets not only enable effective model training but are also crucial for ensuring fairness and generalization in medical AI systems. A range of multimodal datasets has been curated to support various medical imaging and diagnostic tasks.

\begin{table}[h]
    \centering
    \resizebox{\textwidth}{!}{%
        \begin{tabular}{p{4cm} >{\centering\arraybackslash}p{6cm} >{\centering\arraybackslash}p{7cm} >{\centering\arraybackslash}p{4cm}}
            \toprule
            \textbf{Dataset} & \textbf{Modalities} & \textbf{Size} & \textbf{Application} \\
            \midrule
            \textbf{2D-image-text} \\
            \midrule
            CheXpert \cite{irvin2019chexpert} & X-ray, Reports, Labels & 224K triplets & Chest X-ray \\
            CheXinstruct \cite{chen2024chexagent} & X-ray, Instructions & 8.5M instruction triplets & Chest X-ray \\
            Harvard-FairVLMed \cite{luo2024fairclip} & SLO fundus images, Demographics, Notes & 10K samples & Ophthalmology \\ 
            MedTrinity-25M \cite{xie2024medtrinity} & Medical images, Captions, Bounding-boxes & 25M pairs & Medical imaging \\
            MedVidQA \cite{gupta2023dataset} & Medical videos, Labels, QA-pairs & 6K videos, 6K labels, 3K QA & Medical videos \\
            MIMIC-CXR \cite{johnson2019mimic} & X-ray, Reports & 377K images, 227K reports & Chest X-ray \\
            MS-CXR \cite{boecking2022making} & X-ray, Descriptions, Bounding-boxes & 1K image-sentence pairs, Bounding-boxes & Chest X-ray \\
            OmniMedVQA \cite{hu2024omnimedvqa} & Medical images, QA & 118K images, 127K QA-pairs & Medical imaging \\
            OpenPath \cite{huang2023visual} & Histopathology images, Captions & 208K pairs & Digital pathology \\
            PadChest \cite{bustos2020padchest} & X-ray, Reports & 160K images, 109K texts & Chest X-ray \\
            PathVQA \cite{he2020pathvqa} & Medical images, QA & 5K images, 33K QA & Medical imaging \\
            PMC-15M \cite{zhang2024multimodal} & Medical images, Captions & 15M image-text pairs & Medical imaging \\
            PubMedVision \cite{chen2024huatuogpt} & Medical images, QA & 1.3M QA pairs & Medical imaging \\
            Quilt-1M \cite{ikezogwo2024quilt} & Histopathology images, Captions & 1M pairs & Digital pathology \\
            Rad-ReStruct \cite{pellegrini2023rad} & X-ray, Structured reports & 3720 images, 3597 Reports & Chest X-ray \\
            SLAKE \cite{liu2021slake} & Medical images, QA & 642 images, 14K QA pairs & Medical imaging \\
            UniMed \cite{khattak2024unimed} & Medical images, Captions & 5.3M image-text pairs & Medical imaging \\
            VQA-RAD \cite{lau2018dataset} & Radiology images, Captions & 315 images, 3.5K QA pairs & Radiology \\
            \midrule
            \textbf{3D-volume-text} \\
            \midrule
            AMOS-MM \cite{amosmm} \cite{ji2022amos} & CT, Reports, QA & 2K image-report pairs, 7K QA & Chest, abdomen, pelvis CT \\
            BrainMD \cite{wangenhancing} & MRI, Reports, EHR & 2.5K cases & Brain MRI \\
            BIMCV-R \cite{chen2024bimcv} & CT, Reports & 8K image-report pairs & CT \\
            CT-RATE \cite{hamamci2024foundation} & CT, Reports, Labels & 25K triplets & Chest CT \\
            INSPECT \cite{huang2023inspect} & CT, Reports, EHR, labels & 23K image-report pairs, EHRs & Chest CT \\
            M3D-Data \cite{bai2024m3d} & CT, Captions, QA, Masks & 120K images, 42K captions, 509K QA, 149K masks & CT \\
            RadGenome-Brain MRI \cite{lei2024autorg} & MRI, Reports, Masks & 3.4K image-region-report triplets & Brain MRI \\
            RadGenome-Chest CT \cite{zhang2024radgenome} & CT, Reports, Masks, QA & 25K image-report pairs, 665K masks, 1.3M QA & Chest CT \\
            \midrule
            \textbf{Others} \\
            \midrule
            Duke Breast Cancer MRI \cite{saha2018machine} & Genomic, MRI images, Clinical data & 922 subjects & Breast cancer \\
            PTB-XL \cite{wagner2020ptb} & ECG signals, Reports, Labels & 21K triplets & ECG \\
            PubChemSTM \cite{liu2023multi} & Chemical structures, Descriptions & 280K chemical structure–text pairs & Drug design \\            
            SwissProtCLAP \cite{liu2023text} & Protein Sequence, Text & 441K sequence-text pairs & Protein design \\
            \bottomrule
        \end{tabular}%
    }
    \caption{Summary of multimodal datasets used for medical AI, grouped by modality categories. The table lists dataset names, the types of modalities (e.g., text and medical images), dataset sizes, and key applications such as image-text retrieval, report generation, and disease classification. (Abbreviations: QA - question answering.)}
    \label{tab:multimodaldatasets}
\end{table}

A substantial proportion of multimodal datasets focus on pairing vision and text data, as this combination is central to tasks where both visual context and descriptive language are critical for diagnostic interpretation. Notable public datasets like MIMIC-CXR \cite{johnson2019mimic} and CheXpert \cite{irvin2019chexpert} provide rich resources for training 2D vision-language models in radiology. These datasets include not only radiology reports but also disease-specific labels, enabling more comprehensive evaluations. For benchmarking report generation, ReXGradient \cite{zhang2024rexrank}, a private benchmark dataset of 10,000 studies collected across 67 medical sites in the United States, offers diverse coverage and serves as a reliable standard for radiology-specific performance evaluation.

Expanding beyond radiology, datasets like Quilt-1M \cite{ikezogwo2024quilt} have introduced multimodal resources covering additional domains such as digital pathology \cite{ferber2024context, lu2024multimodal}.

Recent advancements have also led to datasets tailored for 3D imaging modalities such as CT \cite{amosmm, chen2024bimcv, hamamci2024foundation, huang2023inspect} and MRI \cite{lei2024autorg}. Notably, RadMD \cite{wu2023towards} integrates both 2D and 3D imaging modalities, supporting a broader range of applications.

In addition to image-text pairs, a few datasets now include task-specific annotations to support specialized applications. For instance, RadGenome-Brain MRI \cite{lei2024autorg} and RadGenome-Chest CT \cite{zhang2024radgenome} provide segmentation masks, while datasets like MedTrinity-25M \cite{xie2024medtrinity} offer bounding box annotations. These annotations are critical for grounding text predictions to specific regions of interest, enhancing both explainability and diagnostic accuracy in multimodal models.

The data formats of multimodal datasets also vary significantly based on their intended use cases. While datasets like OpenPath \cite{huang2023visual} present images from publicly available sources in formats such as JPEG, datasets like MIMIC-CXR \cite{johnson2019mimic} and CT-RATE \cite{hamamci2024foundation} preserve clinical formats such as Digital Imaging and Communications in Medicine (DICOM) and Neuroimaging Informatics Technology Initiative (NIfTI). These formats are essential for maintaining complete clinical information and enabling compatibility with healthcare systems.

Beyond traditional imaging and text combinations, datasets have also begun exploring additional modalities for specialized biomedical tasks. For example, SwissProtCLAP \cite{liu2023text} integrates protein sequence data to support protein design frameworks, highlighting the potential of multimodal datasets to extend AI applications beyond diagnostic imaging into molecular and genomic research.

\section{Evaluation metrics for generative AI in medicine}
Evaluating generative AI in medicine is essential to ensure models produce accurate, clinically relevant, and reliable outputs \cite{ostmeier2024green}. This section explores evaluation metrics for both text generation, such as radiology report generation, and image generation, emphasizing the importance of clinical validity and utility. As general-purpose metrics often fall short in capturing medical accuracy, domain-specific approaches are required.

\begin{figure}[h!]
    \centering
    \resizebox{0.9\textwidth}{!}{
        \includegraphics{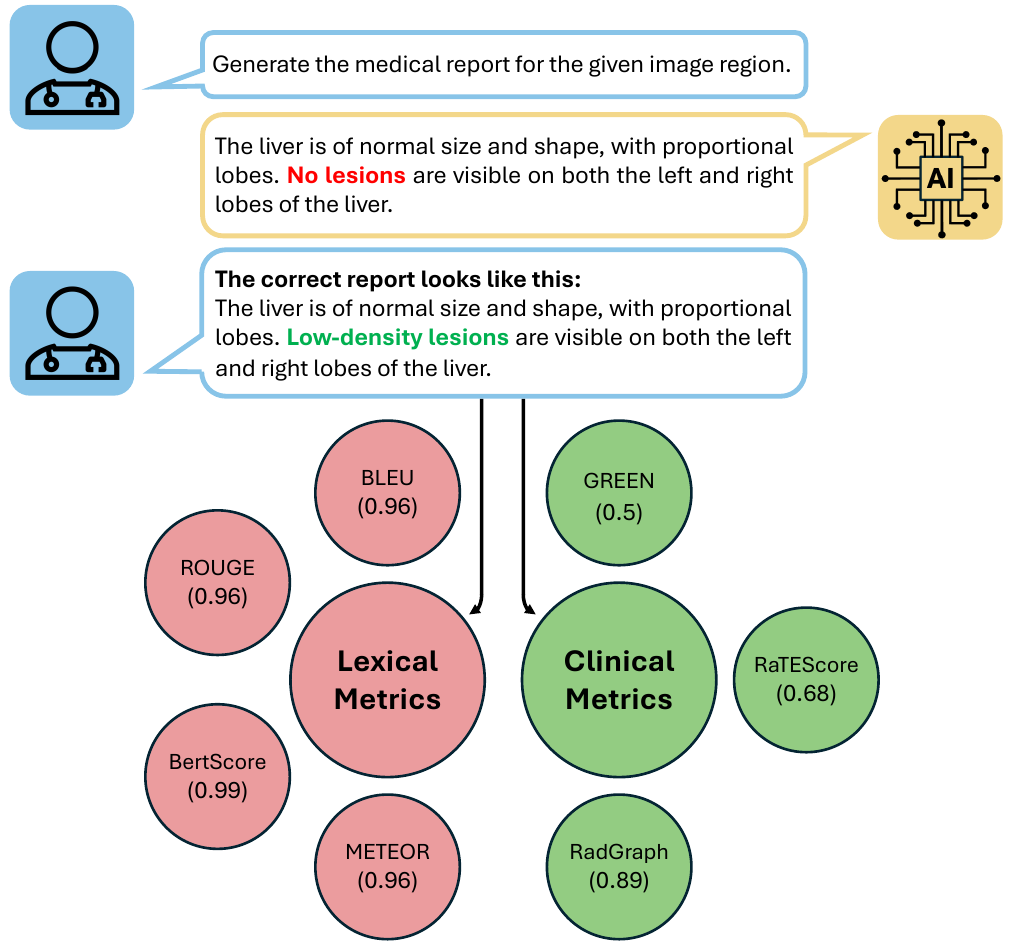}
    }
    \caption{Evaluation of generative AI in medicine: Lexical metrics from the general domain cannot completely capture the clinical correctness as they mainly cover text similarity. Clinically-relevant metrics like GREEN \cite{ostmeier2024green}, RaTEScore \cite{zhao2024ratescore}, or RadGraph \cite{yu2023evaluating} also evaluate the clinical correctness.}
    \label{fig:architectures}
\end{figure}

As report generation is a key application of generative AI in medicine, research has focused on developing reliable evaluation strategies. While standard lexical metrics such as BLEU \cite{papineni2002bleu}, ROUGE \cite{lin2004rouge}, and METEOR \cite{banerjee2005meteor} are commonly used, they often fail to reflect clinical accuracy, as high scores can be achieved despite factually incorrect outputs. To address this, specialized clinical metrics tailored for report generation have emerged (Table \ref{tab:evaluationmetrics}).

For instance, NER-based metrics like RaTEScore \cite{zhao2024ratescore} evaluate key medical entities extracted from both predicted and ground truth reports, offering a more targeted assessment of clinical relevance. RadFact \cite{bannur2024maira} further incorporates grounding by assessing factual correctness against reference image annotations. The GREEN metric described in \cite{ostmeier2024green} goes beyond standard evaluations by integrating error detection with explanations. It provides a clinically grounded score alongside human-interpretable feedback on significant errors, making it a promising tool for both model validation and iterative improvement. ReXrank \cite{zhang2024rexrank}, a benchmark for chest X-ray report generation, combines lexical and clinical metrics for more task-specific assessment.

Additionally, clinical efficacy can be measured using standard classification metrics, such as precision, recall, sensitivity, specificity, and F1-score, particularly when evaluation datasets include labeled disease categories \cite{johnson2019mimic, hamamci2024foundation}. A text classifier can be trained on the generated reports to predict labels, enabling a more structured evaluation of diagnostic accuracy.

\begin{table}[h]
    \centering
    \resizebox{\textwidth}{!}{
        \begin{tabular}{p{3cm} >{\centering\arraybackslash}p{4cm} >{\centering\arraybackslash}p{4cm}}
            \toprule
            \textbf{Metric} & \textbf{Type} & \textbf{Application} \\
            \midrule
            CheXbert \cite{smit2020chexbert} & Classification & Chest X-ray report labeling \\
            CRAFT-MD \cite{johri_evaluation_2025} & Generative & Conversation evaluation \\
            FineRadScore \cite{huang2024fineradscore} & Generative & Report evaluation \\
            GREEN \cite{ostmeier2024green} & Generative & Report evaluation \\
            \citet{ong2024shortcut} & Calibration & Model generalization \\
            RadCliQ \cite{yu2023evaluating} & Composite metric & Report evaluation \\
            RadFact \cite{bannur2024maira} & Grounding & Grounded report evaluation \\
            RadGraph-F1 \cite{yu2023evaluating} & NER similarity & Report evaluation \\
            RaTEScore \cite{zhao2024ratescore} & NER similarity & Report evaluation \\
            \bottomrule
        \end{tabular}
    }
    \caption{Evaluation metrics for medical report generation. This table summarizes key metrics used to evaluate generative AI systems in medical report generation, categorized by type and primary application. (Abbreviations: NER - named entity recognition)}
    \label{tab:evaluationmetrics}
\end{table}

Evaluating image generation in medical AI requires considerations beyond standard image quality metrics like Fréchet Inception Distance \cite{heusel2017gans} and mean squared error. Since synthetic medical images are often used for data augmentation or diagnostic training, their clinical utility must be assessed alongside visual quality. One effective strategy involves generating condition-specific medical images and training a classifier on the synthetic data to evaluate its generalization performance on real clinical cases \cite{hamamci2023generatect}. This ensures that the generated images are not only visually realistic but also contribute to model performance on downstream tasks, such as disease classification and segmentation.

Despite advancements in specialized evaluation metrics for both text and image generation, challenges remain regarding their generalizability across clinical sites and datasets. Frameworks like ReXamine-Global \cite{banerjee2024rexamine} address this by evaluating the robustness of metrics across diverse institutions and data distributions. For text generation, a combination of lexical metrics and clinically grounded assessments is essential to ensure factual correctness and clinical relevance. Similarly, for image generation, both visual quality and downstream clinical utility, such as diagnostic performance on real clinical cases, should be jointly evaluated. Ultimately, a multi-dimensional evaluation approach that considers both data diversity and task-specific requirements is crucial for the safe and effective deployment of generative AI in healthcare.

\section{Discussion}
In this scoping review, we systematically explored the evolution of generative AI in medicine, focusing on LLMs, multimodal LLMs, and their evaluation metrics. Using the PRISMA-ScR framework \cite{tricco2018prisma}, we collected 144 papers published between January 2020 and December 2024 from PubMed, IEEE Xplore, and Web of Science, complemented by a manual search to ensure comprehensive coverage. Our findings highlight the shift from unimodal LLMs focused on textual tasks to more complex multimodal systems capable of integrating medical images, clinical notes, and structured data. These models have shown promise in enhancing diagnostic support, automating clinical workflows, and reducing the workload of healthcare professionals. 

LLMs have advanced biomedical language processing, improving tasks like medical report summarization, named entity recognition, and conversational AI. Adaptation techniques such as supervised finetuning, reinforcement learning, and RAG have further specialized language models for clinical tasks. However, reliance on static datasets like MIMIC-IV \cite{johnson2023mimic} limits the ability to capture evolving medical knowledge. Moreover, privacy issues persist due to the need for extensive data deidentification, and dataset biases can affect fairness by overrepresenting specific populations \cite{wang2023ethical, li2023ethics}.

Multimodal LLMs extend LLM capabilities by integrating multiple data types, such as medical images and text, to address tasks like report generation, cross-modal retrieval, and clinical question answering. Despite these advancements, data heterogeneity remains a challenge, as clinical datasets often vary significantly in format, quality, and completeness across institutions. Additionally, most widely used datasets, such as MIMIC-CXR and CT-RATE \cite{johnson2019mimic, hamamci2024foundation}, focus heavily on radiology, limiting the generalizability of models to other medical domains.

Evaluating generative AI models in medicine requires specialized metrics that go beyond standard language evaluation metrics. While lexical metrics like BLEU \cite{papineni2002bleu} and ROUGE \cite{lin2004rouge} are commonly used, they often fail to capture clinical relevance and factual accuracy. To address this, domain-specific metrics such as RadGraph \cite{yu2023evaluating}, RaTEScore \cite{zhao2024ratescore}, and GREEN \cite{ostmeier2024green} have been developed to assess the clinical validity of generated medical reports. However, challenges remain in standardizing evaluation practices across diverse medical tasks and datasets. Most evaluations are limited to radiology, with less attention given to other specialties. The limited availability of well-annotated multimodal datasets with fine-grained clinical labels further complicates performance benchmarking. Additionally, only a few benchmarking frameworks, such as ReXrank \cite{zhang2024rexrank}, offer the ability to neutrally evaluate models on non-public datasets, limiting comparative performance assessments across different models and data sources. Expanding such benchmarks and ensuring their applicability to a broader range of clinical tasks is essential for developing trustworthy generative models in medicine. 

While this scoping review provides a comprehensive overview of generative AI advancements in medicine, it has certain limitations. Despite the systematic search strategy using the PRISMA-ScR framework, the literature search may not have captured all relevant studies due to the rapidly evolving nature of the field. To mitigate this, a manual search was conducted alongside the database queries to ensure the inclusion of recent and high-impact publications. Moreover, while efforts were made to cover multiple clinical specialties, there remains an overrepresentation of radiology-focused datasets and models, reflecting a broader trend in the literature. We aimed to balance the inclusion of topics and application areas by diversifying the datasets and models included in our analysis, but certain domains such as pathology and genomics remain less represented due to the current availability of multimodal datasets in these fields. 

To further advance the development and responsible deployment of generative AI in medicine, several areas need attention \cite{paschali2024foundation, bluethgen2024best, hager2024evaluation}. First, evaluation frameworks need to evolve beyond lexical metrics by incorporating clinically grounded assessments and domain-specific error analysis. Second, expanding the diversity of training datasets is critical. The current overrepresentation of western institutions and radiology-focused datasets risks introducing biases that limit global applicability \cite{johnson2019mimic, irvin2019chexpert}. Future datasets should encompass a wider range of medical specialties, imaging modalities, and patient demographics, with careful attention to privacy protection and data fairness. Third, improving model explainability remains a priority \cite{lightman2023let, chua2023tackling}. Techniques such as region-specific grounding can help build clinician trust. Finally, the emergence of generalist models \cite{zhang2024generalist, zhou2024generalist} capable of handling multiple modalities and tasks within a unified architecture represents an important step forward, but broader coverage across medical specialties and improved datasets remain essential for widespread adoption. \\

This scoping review provides a structured analysis of the evolution from unimodal LLMs to multimodal generative AI models in medicine, highlighting their potential for improving diagnostic support, clinical documentation, and decision-making. However, challenges related to data diversity, clinical relevance, model interpretability, and the standardization of evaluation metrics remain critical barriers to widespread adoption. Addressing these challenges through interdisciplinary collaboration, improved datasets, and clinically grounded evaluation strategies will be essential to ensure the responsible deployment of generative AI in healthcare.

\newpage

\section*{Additional information}

\noindent\textbf{Acknowledgements}
This work was partially funded via the EVUK programme (“Next-generation AI for Integrated Diagnostics”) of the Free State of Bavaria, the Deutsche Forschungsgemeinschaft (DFG), and Friedrich-Alexander-Universität Erlangen-Nürnberg within the funding program Open Access Publication Funding.
\\

\noindent\textbf{Author contributions} The idea for this review article was developed by all authors. L.B. performed the literature search, paper screening, and selection. The first draft of the manuscript was written by L.B. and subsequently refined by L.B. and S.T.A.. S.T.A. provided clinical expertise. L.B., M.K., N.N., A.M., and S.T.A. provided technical expertise. All authors revised the manuscript and approved the final version for submission. \\

\noindent\textbf{Competing interests} The authors declare no competing interests. \\

\noindent\textbf{Ethical approval} No human or animal subjects are involved in this study. \\

\noindent\textbf{Consent to participate} No human or animal subjects are involved in this study. \\

\noindent\textbf{Consent to publish} No human or animal subjects are involved in this study.
\\

\noindent\textbf{List of Abbreviations}
\begin{itemize}
    \item AI - Artificial intelligence
    \item API - Application programming interface
    \item CLIP - Contrastive language-image pretraining
    \item CoT - Chain-of-thought
    \item CT - Computed tomography
    \item DICOM - Digital Imaging and Communications in Medicine
    \item ECG - Electrocardiogram
    \item EHR - Electronic health record
    \item LLM - Large language model
    \item MLLM - Multimodal large language models
    \item MRI - Magnetic resonance imaging
    \item NER - Named entity recognition
    \item NIfTI - Neuroimaging Informatics Technology Initiative
    \item PRISMA - Preferred Reporting Items for Systematic Reviews and Meta-Analyses
    \item PRISMA-ScR - Preferred Reporting Items for Systematic reviews and Meta-Analyses extension for Scoping Reviews
    \item QA - Question answering
    \item RAG - Retrieval augmented generation
    \item RLHF - Reinforcement learning from human feedback
    \item RLAIF - Reinforcement learning from AI feedback
    \item SFT - Supervised finetuning
\end{itemize}

\newpage
\bibliography{main}

\newpage
\appendix
\renewcommand{\thetable}{S.\arabic{table}}
\setcounter{table}{0} 

\section{Supplementary information}

\subsection{PRISMA-ScR Checklist}\label{checklist}

\begin{longtable}{p{3cm} p{1cm} p{6cm} p{1.5cm}}
\toprule
\textbf{Section} & \textbf{Item} & \textbf{PRISMA-ScR checklist item} & \textbf{Section} \\
\midrule
\endfirsthead

\toprule
\textbf{Section} & \textbf{Item} & \textbf{PRISMA-ScR checklist item} & \textbf{Section} \\
\midrule
\endhead

\endfoot

\textbf{Title} &  &  & \\
\addlinespace
\hline
\addlinespace
Title & 1 & Identify the report as a scoping review. & Title \\
\addlinespace
\hline
\addlinespace
\textbf{Abstract} &  &  & \\
\addlinespace
\hline
\addlinespace
Structured summary & 2 & Provide a structured summary that includes (as applicable): background, objectives, eligibility criteria, sources of evidence, charting methods, results, and conclusions that relate to the review questions and objectives. & Abstract \\
\addlinespace
\hline
\addlinespace
\textbf{Introduction} &  &  & \\
\addlinespace
\hline
\addlinespace
Rationale & 3 & Describe the rationale for the review in the context of what is already known. Explain why the review questions/objectives lend themselves to a scoping review approach. & 1 \\
\addlinespace
Objectives & 4 & Provide an explicit statement of the questions and objectives being addressed with reference to their key elements (e.g., population or participants, concepts, and context) or other relevant key elements used to conceptualize the review questions and/or objectives. & 1 \\
\addlinespace
\hline
\addlinespace
\textbf{Methods} &  &  & \\
\addlinespace
\hline
\addlinespace
Protocol and registration & 5 & Indicate whether a review protocol exists; state if and where it can be accessed (e.g., a Web address); and if available, provide registration information, including the registration number. & 2 \\
\addlinespace
Eligibility criteria & 6 & Specify characteristics of the sources of evidence used as eligibility criteria (e.g., years considered, language, and publication status), and provide a rationale. & 2.1 \\
\addlinespace
Information sources & 7 & Describe all information sources in the search (e.g., databases with dates of coverage and contact with authors to identify additional sources), as well as the date the most recent search was executed. & 2.2 \\
\addlinespace
Search & 8 & Present the full electronic search strategy for at least 1 database, including any limits used, such that it could be repeated. & A.2 \\
\addlinespace
Selection of sources of evidence & 9 & State the process for selecting sources of evidence (i.e., screening and eligibility) included in the scoping review. & 2.3 \\
\addlinespace
Data charting process & 10 & Describe the methods of charting data from the included sources of evidence (e.g., calibrated forms or forms that have been tested by the team before their use, and whether data charting was done independently or in duplicate) and any processes for obtaining and confirming data from investigators. & 2.4 \\
\addlinespace
Data items & 11 & List and define all variables for which data were sought and any assumptions and simplifications made. & 2.3 \\
\addlinespace
Critical appraisal of individual sources of evidence & 12 & If done, provide a rationale for conducting a critical appraisal of included sources of evidence; describe the methods used and how this information was used in any data synthesis (if appropriate). & N/A \\
\addlinespace
Synthesis of results & 13 & Describe the methods of handling and summarizing the data that were charted. & 2.5 \\
\addlinespace
\hline
\addlinespace
\textbf{Results} &  &  & \\
\addlinespace
\hline
\addlinespace
Selection of sources of evidence & 14 & Give numbers of sources of evidence screened, assessed for eligibility, and included in the review, with reasons for exclusions at each stage, ideally using a flow diagram. & 3 \\
\addlinespace
Characteristics of sources of evidence & 15 & For each source of evidence, present characteristics for which data were charted and provide the citations & 4, 5, 6 \\
\addlinespace
Critical appraisal within sources of evidence & 16 & If done, present data on critical appraisal of included sources of evidence (see item 12). & N/A \\
\addlinespace
Results of individual sources of evidence & 17 & For each included source of evidence, present the relevant data that were charted that relate to the review questions and objectives. & 4, 5, 6 \\
\addlinespace
Synthesis of results & 18 & Summarize and/or present the charting results as they relate to the review questions and objectives & 4, 5, 6 \\
\addlinespace
\hline
\addlinespace
\textbf{Discussion} &  &  & \\
\addlinespace
\hline
\addlinespace
Summary of evidence & 19 & Summarize the main results (including an overview of concepts, themes, and types of evidence available), link to the review questions and objectives, and consider the relevance to key groups. & 7 \\
\addlinespace
Limitations & 20 & Discuss the limitations of the scoping review process & 7 \\
\addlinespace
Conclusions & 21 & Provide a general interpretation of the results with respect to the review questions and objectives, as well as potential implications and/or next steps. & 7 \\
\addlinespace
\hline
\addlinespace
\textbf{Funding} &  &  & \\
\addlinespace
\hline
\addlinespace
Funding & 19 & Describe sources of funding for the included sources of evidence, as well as sources of funding for the scoping review. Describe the role of the funders of the scoping review. & Additional information \\
\addlinespace

\bottomrule
\addlinespace
\caption{PRISMA-ScR checklist \cite{tricco2018prisma}} \\
\label{tab:checklist}
\end{longtable}

\newpage
\subsection{Database search queries}\label{queries}

\begin{table}[h]
    \centering
    \resizebox{\textwidth}{!}{
    \begin{tabular}{p{2.5cm} p{1.5cm} p{10cm}}
        \toprule
        \textbf{Database} & \textbf{Results} & \textbf{Search string} \\
        \midrule
        PubMed & 1,325 & ("medic*"[TIAB] OR "healthcare"[TIAB] OR "clinic*"[TIAB] OR "diagnosis*"[TIAB] OR "biomedical"[TIAB]) AND ("language model*"[TIAB] OR "LLM"[TIAB]) NOT ("ChatGPT"[TIAB]) NOT (review[PT]) \\
         & 390 & ("medic*"[TIAB] OR "healthcare"[TIAB] OR "clinic*"[TIAB] OR "diagnosis*"[TIAB] OR "biomedical"[TIAB]) AND ("language"[TIAB] OR "language model*"[TIAB] OR "LLM"[TIAB]) AND ("multimodal"[TIAB] OR "multi-modal"[TIAB] OR "generalist"[TIAB] OR "CLIP*"[TIAB]) NOT ("ChatGPT"[TIAB]) NOT (review[PT]) \\
        \hline
        \addlinespace
        IEEE Xplore & 893 & ("All Metadata":"medic*" OR "All Metadata":"healthcare" OR "All Metadata":"clinic*" OR "All Metadata":"diagnosis*" OR "biomedical") AND ("All Metadata":"language model*" OR "All Metadata":"LLM") NOT ("All Metadata":"ChatGPT") NOT ("All Metadata":"Review") NOT ("All Metadata":"Study") \\
         & 240 & ("All Metadata":"medic*" OR "All Metadata":"healthcare" OR "All Metadata":"clinic*" OR "All Metadata":"diagnosis*" OR "biomedical") AND ("All Metadata":"language" OR "All Metadata":"language model*" OR "LLM") AND ("All Metadata":"multimodal" OR "All Metadata":"multi-modal" OR "generalist" OR "CLIP*") NOT ("All Metadata":"ChatGPT") NOT ("All Metadata":"Review") NOT ("All Metadata":"Study") \\
        \hline
        \addlinespace
        Web of Science & 1,111 & TS=("medic*" OR "healthcare" OR "clinic*" OR "diagnosis*" OR "biomedical") AND TS=("language model*" OR "LLM") NOT TS=("ChatGPT") AND DT=("Article") \\
         & 425 & TS=("medic*" OR "healthcare" OR "clinic*" OR "diagnosis*" OR "biomedical") AND TS=("language" OR "language model*" OR "LLM") AND TS=("multimodal" OR "multi-modal" OR "generalist" OR "CLIP*") NOT TS=("ChatGPT") AND DT=("Article") \\
        \bottomrule
    \end{tabular}
    }
    \caption{Search results from different databases}
    \label{tab:queries}
\end{table}

\end{document}